\documentclass[a4paper,twoside]{article}

\usepackage{epsfig}
\usepackage{subcaption}
\usepackage{calc}
\usepackage{amssymb}
\usepackage{amstext}
\usepackage{amsmath}
\usepackage{amsthm}
\usepackage{multicol}
\usepackage{pslatex}
\usepackage{apalike}
\usepackage{algorithm2e}
\usepackage[bottom]{footmisc}
\usepackage{SCITEPRESS}     % Please add other packages that you may need BEFORE the SCITEPRESS.sty package.

\usepackage{hyperref}

\begin{document}
\title{ODPG: Outfitting Diffusion with Pose Guided Condition}

\author{
    \authorname{
        Seohyun Lee\textsuperscript{*}, 
        Jintae Park\textsuperscript{*}, 
        and Sanghyeok Park\textsuperscript{*}
    }
    \affiliation{Korea University, Seoul, South Korea}
    \email{\{happy8825, sonjt00, huanhuan1235\}@korea.ac.kr}
    \vspace{0.3em} 
    \textsuperscript{*}\scriptsize{These authors contributed equally.}
}


\abstract{Virtual Try-On (VTON) technology allows users to visualize how clothes would look on them without physically trying them on, gaining traction with the rise of digitalization and online shopping. Traditional VTON methods, often using Generative Adversarial Networks (GANs) and Diffusion models, face challenges in achieving high realism and handling dynamic poses. This paper introduces Outfitting Diffusion with Pose Guided Condition (ODPG), a novel approach that leverages a latent diffusion model with multiple conditioning inputs during the denoising process. By transforming garment, pose, and appearance images into latent features and integrating these features in a UNet-based denoising model, ODPG achieves non-explicit synthesis of garments on dynamically posed human images. Our experiments on the FashionTryOn and a subset of the DeepFashion dataset demonstrate that ODPG generates realistic VTON images with fine-grained texture details across various poses, utilizing an end-to-end architecture without the need for explicit garment warping processes. Future work will focus on generating VTON outputs in video format and on applying our attention mechanism, as detailed in the Method section, to other domains with limited data.}

\onecolumn \maketitle \normalsize \setcounter{footnote}{0} \vfill

\section{Introduction}
Virtual Try-On (VTON) technology allows users to visualise how clothes would look on them without physically trying them on. Recently, VTON technology has seen a significant surge in demand, driven by advancements in digitalization and the growing preference for online shopping experiences. The VTON process involves sophisticated image synthesis techniques that map clothing items onto images of users, creating a realistic visual representation.

Some previous popular approaches in VTON utilized Generative Adversarial Networks (GANs), with a generator network creating realistic clothing images on users and a discriminator network evaluating the authenticity of these images \cite{goodfellow2014}. The generator improves over time by learning to produce images that are indistinguishable by the discriminator, enabling the generation of high-quality and realistic try-on images. However, the use of GANs in VTON faces significant challenges. One major issue is the mode collapse problem, where samples generated by GANs lack diversity even when trained on multimodal data. This leads to critical issue in VTON task as GANs continuously generate similar results. Additionally, the adversarial nature of GAN training can lead to unstable synthesized results, as the generator and discriminator oscillate rather than converge to a stable point. This instability is exacerbated when the discriminator becomes too powerful, causing vanishing gradients and halting the generator's updates.

Therefore, recent VTON works prefer diffusion-based methods \cite{gou2023} \cite{kim2023}, which synthesize more realistic images through a series of denoising steps. Stable Diffusion (SD) \cite{lu2024} \cite{rombach2022} is the most frequently used method in the diffusion-based VTON domain. However, there seems to be a constraint when using Stable Diffusion directly for VTON applications. Stable Diffusion generates images through a process of denoising guided by text information. Due to the difference in information densities between language and vision, the produced images often exhibit ambiguity. Textual descriptions may not capture the fine details and complexities of visual appearances, leading to images that do not accurately reflect the intended appearance. 

To address these challenges, we introduce Outfitting Diffusion with Pose Guided Condition (ODPG) method that first transforms the target garment image into latent features using a VAE and Swin Transformer. We then incorporate these features as conditions in the denoising process, enabling non-explicit synthesis of the garment to source human image. In this way, we generate VTON images based on dynamic poses by incorporating additional clothing latent conditions unlike traditional VTON models which often produce images with reduced realism and are limited to specific static poses. 

The main contributions of our ODPG method are:
\begin{itemize}
\item \textbf{Multi-Image-Conditioned Controllable Generation for VTON:} We introduce a novel model for VTON, designed to capture a rich semantic understanding of the target garment, target pose, and source appearance. Our architecture enhances the generation process by embedding a target garment image as an additional condition in the denoising process.

\item \textbf{VTON Without Explicit Garment Warping:} Unlike traditional VTON models that rely on explicit garment warping processes, our approach integrates garment information implicitly throughout the denoising process. This allows for seamless garment alignment with various human poses and appearances without the need for warping.

\item \textbf{Bias-Augmented Query Attention Mechanism:} We propose a bias-augmented query attention mechanism that allows the model to learn complex relationships between the source appearance, target pose, and target garment. This mechanism enhances the model’s ability to generalize to new combinations of appearances, poses, and garments, thereby reducing the need for a four-paired dataset.

\item \textbf{End-to-End Dynamic VTON Model:} We present a novel end-to-end VTON model that incorporates pose transfer in a single unified training process. Unlike existing approaches that rely on multi-stage pipelines, this architecture simplifies the training process while maintaining high-quality synthesis results.

\end{itemize}

\section{Related Work}
\subsection{Controllable Diffusion Models}

Given the challenges with GAN-based methods \cite{goodfellow2014}, which struggle with instability and difficulty in generating high-quality images in a single pass, diffusion models \cite{ho2020} have become a preferred alternative for high-resolution image synthesis. These models work by progressively creating realistic images through multiple denoising steps. Advances in controllable diffusion models, such as Stable Diffusion, have introduced additional control signals during the data generation process. This innovation allows for more accurate and targeted data generation, enabling users to control specific attributes of the output. This increased flexibility and utility extend to various applications, including the VTON domain. 

Therefore, we aim to achieve realistic and controllable VTON results by incorporating the source human image, target garment image, and target pose image as conditions into the denoising stages of the controllable diffusion model architecture.

\subsection{Pose-Guided Person Image Synthesis}

Pose-Guided Person Image Synthesis is a technique where a new image of a person in a different pose is generated from an existing image. This task was initially introduced by Ma et al., who generated new images of individuals in different poses using an adversarial approach to enhance image quality. 

Early techniques focused on separating pose and appearance features by learning pose-irrelevant information; therefore, they struggled with complex textures and detailed appearances. To address this issue, auxiliary information like parsing maps and heatmaps were incorporated, improving generation quality by breaking down images into semantically meaningful parts. Recent advancements have further refined spatial alignment between pose and appearance using diffusion models like PIDM \cite{bhunia2023} and PoCoLD \cite{han}, which mitigate the instability of GANs and enhance high-resolution image synthesis. These advanced methods focus on spatial correspondence through cross-attention mechanisms, though they risk overfitting by merely aligning source appearance to the target pose. 

Our approach, however, is more efficient and end-to-end, freezing most parameters and avoiding reliance on image-text pairs, making it more adaptable for dynamic VTON tasks.

\subsection{Coarse-to-Fine Latent Diffusion for Pose-Guided Person Image Synthesis}

PIDM \cite{bhunia2023} and PoCoLD \cite{han} enhance the alignment between the source image and the target pose to generate more realistic textures, but they struggle to achieve a high-level semantic understanding of human images. This shortfall often results in overfitting and poor generalization, especially when synthesizing exaggerated poses that deviate significantly from the source image or are uncommon in the training data. 

Lu et al. address these issues with a Coarse-to-Fine Latent Diffusion (CFLD) \cite{lu2024} approach, which integrates a perception-refined decoder and a hybrid-granularity attention module to improve semantic understanding and texture detail. 

Perception-refined decoder achieves the perception of the source human image by initializing sets of learnable queries randomly and refining them in the next decoder progressively. The output produced by the decoder is regarded as the coarse-grained prompt for source image description, which focuses on the common semantics possessed across all the different person images such as gender and age. 

Hybrid-Granularity Attention Module effectively encodes multi-scale fine-grained human appearance features as bias terms. These bias terms then augment the coarse-grained prompt generated by the Perception-Refined Decoder. This method helps with the alignment of the source image and the target pose by using only the necessary fine-grained details of the coarse-grained prompt. This enhances the quality of the output images by improving generalization.

Unlike previous methods, this novel architecture does not depend on image-caption pairs, making it more adaptable for Pose-Guided Person Image Synthesis tasks.

\begin{figure*}[t]
\centering
\includegraphics[width=\textwidth]{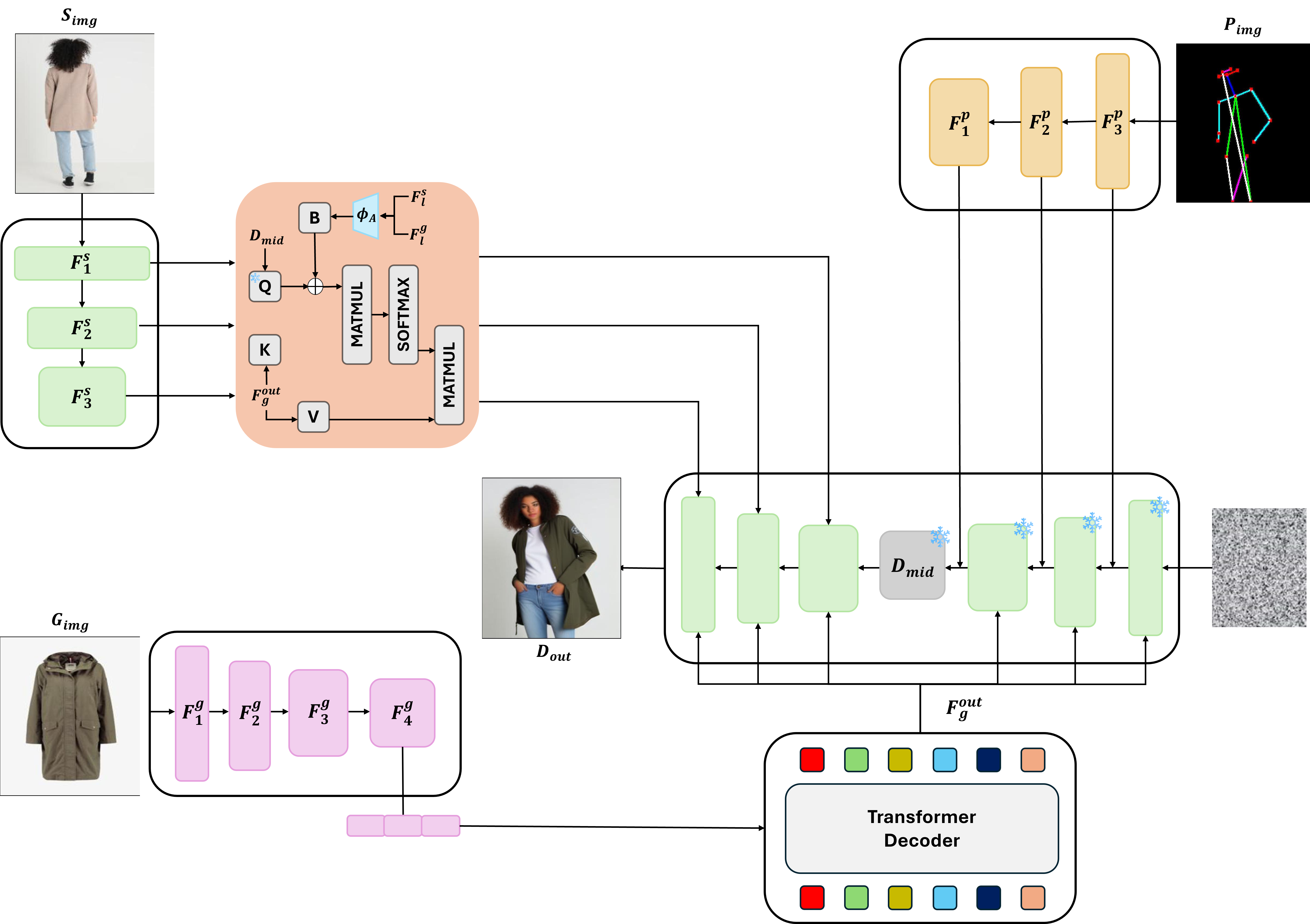}
\caption{Overall Pipeline of ODPG : The source image features are extracted at three different scales using a Swin Transformer. Similarly, the garment image passes through a Swin Transformer, with features extracted at four scales. The fourth-scale feature is passed through a transformer decoder with learnable queries, which outputs a general garment information to condition the diffusion UNet structure at each downsample and upsample block. Multi-scale features from both the source and garment images are combined via attention mechanisms, with fine details added as conditions in the upsampling blocks. Additionally, pose information, processed through a ResNet, provides coarse guidance, conditioning the downsampling blocks. }
\label{fig:odpg_pipeline}
\end{figure*}

\section{Method}

\subsection{Preliminary}
Our approach leverages the principles of latent diffusion models to achieve high-quality image generation. The framework consists of two main components: a Variational Autoencoder (VAE) and a UNet-based denoising model. The VAE is responsible for transforming images from the raw pixel space into a lower-dimensional latent space \cite{Kingma2014AutoEncodingVB}. This transformation facilitates efficient manipulation and processing of the images.

In the diffusion model, the generation process involves a sequence of steps where noise is progressively added to a latent variable, and then gradually removed to produce a clean image. This process is inspired by the Denoising Diffusion Probabilistic Models (DDPM) \cite{ho2020denoising}. The forward diffusion process adds Gaussian noise to the initial latent variable $z_0$ at each timestep $t$, resulting in a noisy latent $z_t$. The relationship is given by:

\begin{equation}
z_t = \sqrt{\bar{\alpha}_t} z_0 + \sqrt{1 - \bar{\alpha}_t} \epsilon,
\end{equation}

where $\epsilon \sim \mathcal{N}(0, I)$ represents the Gaussian noise, and $\bar{\alpha}_t$ is a variance schedule controlling the amount of noise added at each step. The reverse process uses a neural network to predict and subtract the noise, effectively denoising the latent variables to reconstruct the image.

The UNet-based model \cite{ronneberger2015unet} plays a crucial role in this denoising process. It is trained to predict the noise present in the noisy latents $z_t$ conditioned on the timestep $t$. The objective is to minimize the mean squared error between the predicted noise and the actual noise, formalized as:

\begin{equation}
L_{\text{mse}} = \mathbb{E}_{z_0, \epsilon, t} \left[ \| \epsilon - \epsilon_\theta(z_t, t) \|^2 \right].
\end{equation}

\subsection{Our Approach}
\textbf{Method Overview}
As illustrated in Fig \ref{fig:odpg_pipeline}, we introduce a conditioning mechanism that integrates garment, appearance, and pose information into the image synthesis process. In Section \ref{sec:multiscale}, we propose a multi-scale feature extraction approach. Subsequently, in Section \ref{sec:attention}, we present a bias-augmented query attention mechanism to propagate the extracted features through a U-Net-based diffusion model. Finally, in Section \ref{sec:reduce}, we discuss a training strategy designed to improve generalization and reduce dependency on the training data.

\subsubsection{Multi-Scale Feature Extraction}\label{sec:multiscale}\leavevmode
Building upon CFLD for Pose-Guided Person Image Synthesis, which generates fine-grained pose-transferred human images by adding pose and appearance conditions during the denoising process, we extend this approach to include garment, appearance, and pose conditions in the dynamic VTON process. To achieve this, we first extract multi-scale feature maps from the Source Human, Target Garment, and Target Pose images.
Specifically, both the Source Human image and the Target Garment image are passed through a Swin Transformer, yielding multi-scale feature maps for the source, $F_s = [f_{s1}, f_{s2}, f_{s3}]$, and for the garment, $F_g = [f_{g1}, f_{g2}, f_{g3}, f_{g4}]$. Simultaneously, we extract keypoints from the Target Pose image using OpenPose, a pre-trained pose estimation model, and process them through ResNet blocks to obtain multi-scale pose feature maps, $F_p = [f_{p1}, f_{p2}, f_{p3}]$.

\subsubsection{Bias Augmented Query Attention}\label{sec:attention}\leavevmode
In our UNet-based diffusion model, we use a learnable vector as the initial query $Q_{\text{learn}}$ for the attention mechanism. During the denoising process, this query interacts with keys and values corresponding to the target garment via cross-attention. By incorporating pose, appearance, and garment conditions as biases to the query $Q_{\text{learn}}$ and applying the attention output to the noisy latent, the model effectively generates natural garment features that align with the source human and target pose.

To initiate this process, we first flatten the coarsest garment feature map $F_g^4$ and pass it through a Transformer decoder. This step generates $ F_g^{\text{out}}$ which captures general garment information. Based on this feature map, we then compute the keys $K$ and values $V$, which are used throughout the UNet’s denoising process:

\begin{equation}
K = W_k^l F_g^{\text{out}}, \quad V = W_v^l F_g^{\text{out}}
\end{equation}

Here, $W_k^l$ and $W_v^l$ are learnable projection matrices corresponding to the $l$-th scale of denoising blocks. 

Through cross-attention between the query $Q_{\text{learn}}$ and these keys $K$ and values $V$, we aim to extract garment features that correspond to the target pose. 
To achieve this, multi-scale pose feature maps $[f_{p1}, f_{p2}, f_{p3}]$ are added as biases to the query $Q_{\text{learn}}$ within the downsampling blocks. Consequently, the samples processed through these layers effectively incorporate both the pose and garment information, ensuring that the generated garment aligns naturally with the target pose.

In the upsampling blocks, we simultaneously integrate garment and appearance information by utilizing the appearance encoder $\phi_A$, as proposed in CFLD. The fine-grained appearance encoder $\phi_A$ is designed to capture detailed texture information and consists of multiple Transformer layers, including a zero convolution layer at both the beginning and end to stabilize training. In our method, garment features $F_g^l$ and appearance features $F_s^l$, extracted at each scale, are passed through the appearance encoder $\phi_A$ to calculate the corresponding biases:

\begin{equation}
B = \phi_A(F_s^l,F_g^l)
\end{equation}

These biases are then incorporated into the query $Q_{\text{learn}}$, adjusting it to reflect appearance and garment information. The overall attention mechanism can be summarized by the following equation:

\begin{equation}
F_o^l = \text{softmax}\left( \frac{(Q_{\text{learn}} + B)K^T}{\sqrt{d}} \right)V.
\end{equation}
Here, The attention output $F_o^l$ from the $l$th layer is then added to the noisy latent, guiding the denoising process at each step. 
Through the entire denoising process described above, the VTON sample, having undergone all the upsampling and downsampling blocks, effectively integrates the source appearance, target garment, and target pose information. This integration results in a realistic synthesized image. By utilizing appearance and pose information as the query to extract corresponding garment features, the VTON output incorporates natural clothing details, such as wrinkles and fit, that emerge when a person of a specific body shape wears the garment and takes on a specific pose.

Moreover, this attention mechanism not only enhances the realism of the synthesized images but also reduces the model's dependency on four-paired datasets. In the following section, we explore how this approach allows us to train the model effectively using existing three-paired datasets, thereby simplifying data requirements for dynamic VTON architectures.

\subsubsection{Enhancing Generalization and Reducing Data Dependency}\label{sec:reduce}\leavevmode
Training a VTON model that simultaneously transforms both garment and pose typically requires a four-paired dataset, including the source human image, target garment image, target pose image, and a ground truth image showing the source person wearing the target garment in the target pose. However, existing datasets, such as DeepFashion \cite{liu2016deepfashion} and FashionTryOn \cite{zheng2019virtually}, are three-paired datasets. In these datasets, the source human, pose-transferred human, and target garment all feature the same garment, which may appear to limit the dynamic VTON model’s ability to generalize to new garments. Previous methods focus on either pose transfer or garment transfer, and thus lack comprehensive pairing for transformations in both aspects.

Our Bias Augmented Query Attention mechanism enables the model to learn complex relationships between the source appearance, target pose, and target garment, enhancing its ability to generalize to new combinations of appearances, poses, and garments, thereby reducing the need for a four-paired dataset. Specifically, the attention mechanism integrates the source appearance and target pose as biases into the learnable query $Q_{\text{learn}}$, guiding the model to extract relevant garment features during the denoising process. This approach allows the model to synthesize realistic images of the source person wearing different, previously unseen garments in new poses, without requiring a ground truth image that combines all transformations.

We acknowledge that in our training setup, the source person is wearing the target garment, which may seem to limit the model's ability to generalize to unseen garments. However, our experimental results demonstrate that the model does not merely replicate the input; instead, it effectively transfers and integrates garment features based on the conditioned appearance and pose without memorizing specific garment instances.

To validate the generalization capability of our model, we tested it with source person images wearing garments different from the target garment and in various poses. The model successfully synthesized realistic images of the source person wearing new, previously unseen garments in new poses. This indicates that our Bias Augmented Query Attention mechanism enables the model to focus on the interactions between garments, poses, and appearances, effectively generalizing beyond the training data.

This approach reduces the dependency on four-paired datasets, allowing us to leverage existing three-paired datasets such as DeepFashion \cite{liu2016deepfashion} and FashionTryOn \cite{zheng2019virtually}, which contain only three components: the source human image, the garment image, and the target posed human image with the same garment.

In summary, our attention mechanism enhances the model's ability to generalize to new combinations of appearances, poses, and garments without relying on exhaustive paired datasets. This alleviates the data requirements for training dynamic VTON architectures.

\section{Optimization}
To optimize the training process, we utilize a self-reconstruction mechanism between the source image and itself. The reconstruction loss is formulated as:

\[
L_{\text{rec}} = \mathbb{E}_{z_0, x_s, x_{sp}, X_{sg}, \epsilon, t} \left[ \| \epsilon - \epsilon_\theta(z_t, t, x_s, x_{sp}, X_{sg}) \|_2^2 \right],
\]

where $z_0 = E(x_s)$ denotes the latent representation of the source image, and $z_t$ is the perturbed latent state produced from $z_0$ at timestep $t$.
 The total loss function is defined as:

\[
L_{\text{overall}} = L_{\text{mse}} + L_{\text{rec}}.
\]

To further enhance the training process, we adopt the cumulative classifier-free guidance to strengthen the source appearance, target pose features and garment features simultaneously. Following the methodology presented in InstructPix2Pix \cite{InstructPix2Pix}, where both image and textual conditionings are randomly dropped during training to allow for conditional or unconditional denoising, we employ a similar mechanism with our multi-conditional inputs(pose, appearance, and garment. By doing so, we ensure flexibility and control in guiding the model's output based on each feature. The modified classifier-free guidance is expressed as:

\begin{align}
\epsilon_t = &\epsilon_\theta(z_t, t, \emptyset, \emptyset) \notag \\
&+ w_{\text{pose}} (\epsilon_\theta(z_t, t, \emptyset, x_{tp}) - \epsilon_\theta(z_t, t, \emptyset, \emptyset)) \notag \\
&+ w_{\text{app}} (\epsilon_\theta(z_t, t, x_s, x_{tp}) - \epsilon_\theta(z_t, t, \emptyset, x_{tp})) \notag \\
&+ w_{\text{garment}} (\epsilon_\theta(z_t, t, x_g, x_{tp}) - \epsilon_\theta(z_t, t, \emptyset, x_{tp})),
\end{align}

where $w_{\text{pose}}$, $w_{\text{app}}$, and $w_{\text{garment}}$ are weighting factors for the pose, appearance, and garment guidance, respectively.
Based on this noise prediction, we utilized the DDIM Scheduler in the sampling process, performing 50 denoising steps. As a result, the synthesized images were successfully based on the embedded conditions of the Source Human Image, Target Pose Image, and Target Garment Image.

\begin{figure*}[t]
\centering
\includegraphics[width=\textwidth]{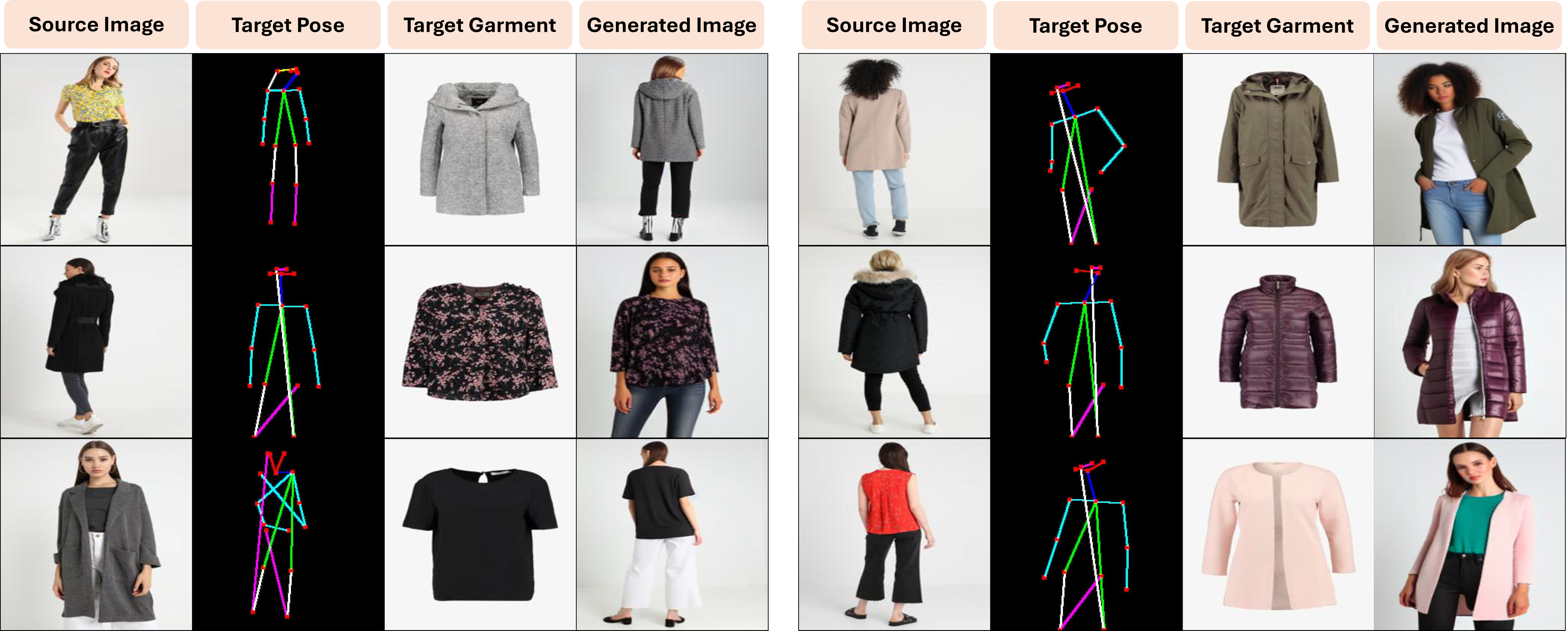}
\caption{Qualitative results showcasing the performance of our model. The images illustrate the ability of our model to retrieve relevant items across different poses and domains.}
\label{fig:qualitative_result}
\end{figure*}

\section{Experiment}
\subsection{Dataset Description}
 Our experiment was conducted on the large-scale In-Shop Clothes Retrieval benchmark of DeepFashion \cite{liu2016deepfashion}. The original DeepFashion database contains over 300,000 cross-pose and cross-domain image pairs. However, since our model's goal is to synthesize VTON images based on new poses and garments, we extracted a dataset consisting of source human images, target pose human images, and target garment images. Furthermore, we utilized only 256×176 low-resolution images to verify that our model is capable of generating fine details even under limited resolution and constrained computational power. Consequently, the final extracted DeepFashion dataset consists of 6,014 images. As this dataset alone is insufficient to fully validate our model's performance, we also utilized the FashionTryOn \cite{zheng2019virtually}. This dataset consists of 21,724 images, including garment images and human model images in different poses, with a resolution of 192×256. We extracted joint keypoints from the model images using OpenPose and subsequently trained our three-image-conditioned architecture using both the FashionTryOn and the extracted DeepFashion datasets.

\subsection{Implementation Details}
We developed our method using the PyTorch framework, incorporating HuggingFace Diffusers based on Stable Diffusion v1.5. For preprocessing, the source, garment images, and pose images are resized to dimensions of 256×256 pixels.

Training was performed for 30 epochs with the Adam optimizer. The training process utilized a single NVIDIA A100 GPU, running for approximately 30 hours, with a batch size configured to 24.

The key aspects of our model configuration are as follows:
\begin{itemize}
\item \textbf{UNet Configuration}: We use a pretrained UNet model, focusing on training specific blocks (indexed from 3 to 11). Cross-attention mechanisms are selectively applied, with only the key and value layers being trainable.
\item \textbf{Pose Guidance}: Pose information is downscaled by a factor of 4, with defined channels for both pose and input data.
\item \textbf{Appearance Guidance}: Convolutional layers are configured with kernel and stride sizes of 1, and attention residual blocks are indexed from 3 to 11 to effectively capture detailed features.
\item \textbf{Classifier-Free Guidance}: During training, conditions are dropped with a probability of 20\%.
\item \textbf{Garment Loss Weight}: The garment loss weight is set to 1.0 to ensure balanced training between the source and garment features.
\end{itemize}

\subsection{Qualitative Analysis}
As illustrated in Fig \ref{fig:qualitative_result}, we attempted to implicitly synthesize the Source Human (col 1) with the target pose (col 2) and target garment (col 3) through the denoising UNet. 

The synthesized results in the output column (col 4) of each left and right table demonstrate that our end-to-end architecture successfully integrates garment and pose information into multiple dynamic-posed VTON results. 

The garment features effectively deliver detailed patterns, wrinkles, textures, and colors of the target garment into the overall denoising process. Additionally, the pose features encoded by several ResNet blocks are successfully reflected in all samples.

\subsection{Quantitative Analysis}
We evaluated our model using FID, SSIM, and LPIPS to compare its performance against other models. These metrics are well-suited to assess the overall quality, structural similarity, and perceptual diversity of generated images. However, given that our model is an end-to-end multi-conditioning approach capable of simultaneously altering both pose and garment, there is an inherent trade-off in metrics when compared to models that either focus solely on garment manipulation or adopt multi-stage processes for pose and garment modifications. 

The results of our experiments, as presented in Table \ref{table: quantitative}, provide a comparison between our multi conditioning end-to-end model (ODPG) and two different types of models on the FashionTryOn dataset: those that condition only on the garment and those that use multi-stage training to modify both the pose and garment.

Garment-only models such as VITON-HD \cite{vitonhd}, GP-VTON \cite{gpvton}, and CP-VTON \cite{cpvton} performed well in metrics like SSIM and LPIPS, since they only focus on conditioning the garment. Our model, however, handles the more complex task of changing both the garment and the pose in a single end-to-end process. Despite this, ODPG achieved reasonable SSIM and LPIPS scores in comparison. Notably, our model outperformed all other models in FID, which indicates that the generated images from ODPG are closer to real images in terms of overall quality and distribution.

The second set of models, such as FashionTryOn \cite{fashiontryon}, Wang et al. \cite{Wang}, and style-VTON \cite{stylevton}, rely on multi-stage training, where the pose and garment are handled in separate steps. FashionTryOn utilizes a bidirectional GAN in multiple stages to enhance both pose and garment deformation, while Wang et al. employ distinct training steps for parsing transformation, spatial alignment, and face refinement. Similarly, style-VTON uses a two-step approach, where a spatial transformation network first aligns the garment, followed by a UNet to refine details.

In contrast, our model’s end-to-end nature simplifies the pipeline while minimizing trade-off in performance. Although the SSIM and LPIPS metrics imply slightly lower performance of  our model compared to some multi-stage models, ODPG demonstrated competitive performance, especially excelling in FID. This suggests that our model effectively captures the visual realism and diversity of the generated images while maintaining the benefits of an end-to-end structure, without the need for separate stages.

\begin{table}[h]
    \centering
    \resizebox{0.485\textwidth}{!}{
    \begin{tabular}{|l|c|c|c|}
        \hline
        \textbf{Method} & \textbf{SSIM $\uparrow$} & \textbf{LPIPS $\downarrow$} & \textbf{FID $\downarrow$} \\ 
        \hline
        \multicolumn{4}{|l|}{\textbf{Pose unchanged (garment only changed)}} \\
        \hline
        VITON-HD & \textbf{0.853} & 0.187 & 44.25 \\ 
        GP-VTON & 0.724 & 0.384 & 66.01 \\ 
        CP-VTON & 0.739 & \textbf{0.159} & 56.23 \\ 
        \hline
        \multicolumn{4}{|l|}{\textbf{Pose and garment both changed}} \\
        \hline
        FashionTryon & 0.699 & 0.211 & 53.911 \\ 
        Wang et al. & 0.695 & 0.152 & 44.306 \\ 
        style-VTON & 0.759 & 0.131 & 39.341 \\ 
        \hline
        \hline
        ODPG (ours, end-to-end) & 0.583 & 0.288 & \textbf{33.625} \\ 
        \hline
    \end{tabular}
    }
    \caption{Quantitative comparisons of two types of models on the FashionTryOn dataset: models that change only the garment, and models that adapt to both pose and garment. The higher the better for SSIM, and the lower the better for LPIPS and FID.}
\label{table: quantitative}
\end{table}

\subsection{Comparison with Other Models}

\begin{figure}[h]
\centering
\includegraphics[width=0.45\textwidth]{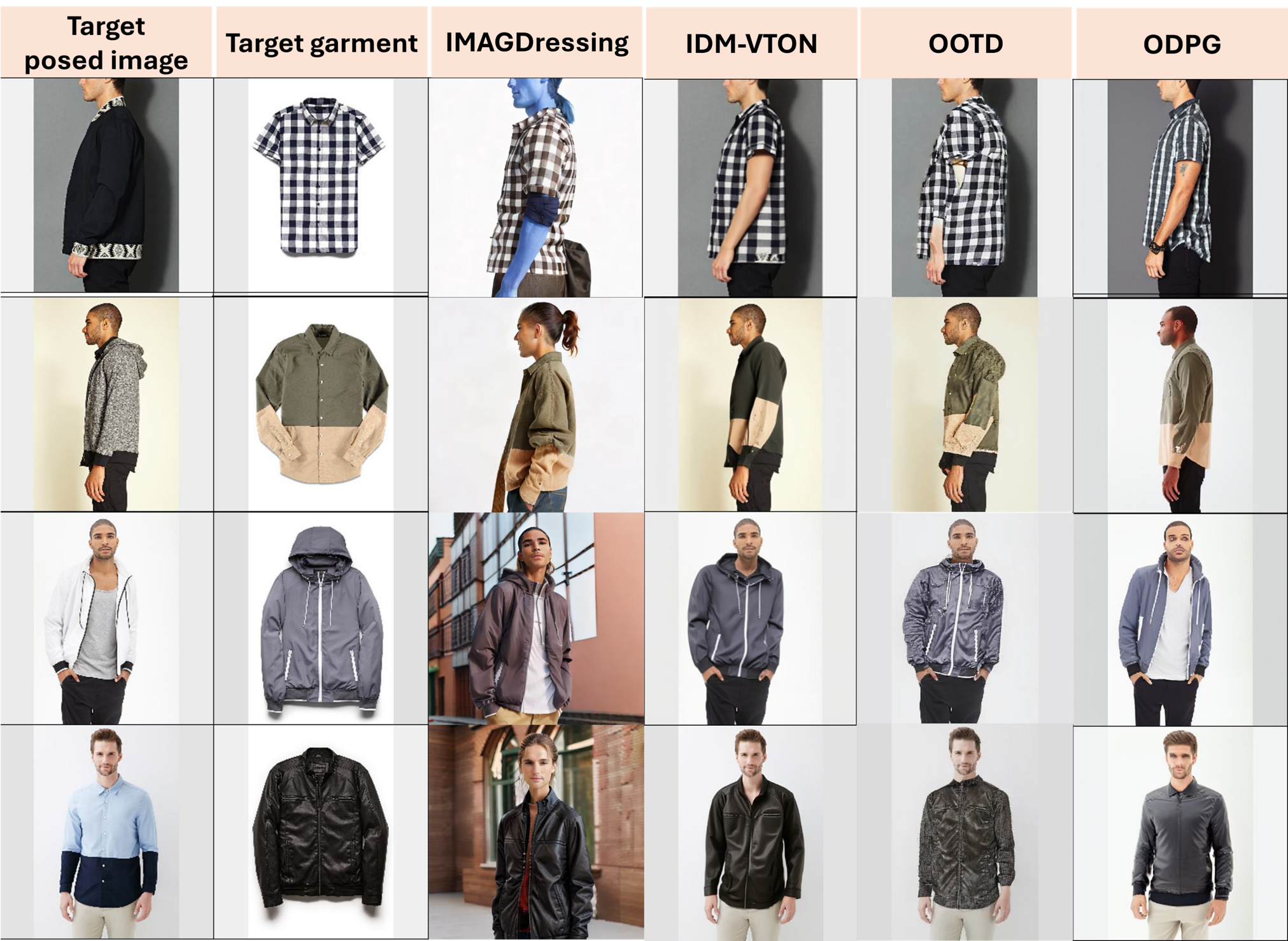}
\caption{Comparison of different models: IMAGDressing, IDM-VTON, OOTD, and our ODPG model.}
\label{fig:comparison}
\end{figure}

As illustrated in Fig \ref{fig:comparison}, we compared our ODPG model with several existing models, including IMAGDressing \cite{IMAGDressing}, IDM-VTON \cite{IDM-VTON}, and OOTDiffusion \cite{OOTD}. A common issue among these previous virtual try-on (VTON) methods is their approach of masking the person and making the garment to fit onto the existing silhouette. This technique often fails to accurately reflect the garment's features and fit, especially when the source image depicts a person wearing thick or layered clothing.

For instance, overlaying a short-sleeved shirt onto a person initially wearing bulky attire can result in the new garment appearing oversized and ill-fitting, mimicking the bulkiness of the original clothing. Similarly, if the person is wearing a hooded sweatshirt and we attempt to dress them in a non-hooded garment, remnants of the hood may awkwardly appear at the back of the neck. These problems arise because the models ambiguously combine the size and shape of the original input garment with the new one, leading to unrealistic distortions and artifacts.

While some models like IMAGDressing attempt to incorporate pose information, they often struggle to preserve the input person's appearance accurately, sometimes altering individual-specific traits or even the perceived gender. Other models such as IDM-VTON and OOTDiffusion maintain the person's characteristics better but still encounter issues like occasional sub-optimal fitting, and some challenges adapting to pose changes.

In contrast, our ODPG model effectively addresses these challenges by first integrating pose and garment information during the downsampling stage and then adding appearance details in the upsampling stage. This approach allows the model to adjust the new garment to the person's body structure, regardless of the original clothing's size or shape. As a result, the new garment achieves a more natural and realistic fit, while maintaining consistency in the overall appearance.

These advancements highlight the strengths of our ODPG model in achieving a more accurate and realistic integration of pose, garment, and appearance features, overcoming the common limitations observed in existing models.

\begin{figure*}[t]
\centering
\includegraphics[width=\textwidth]{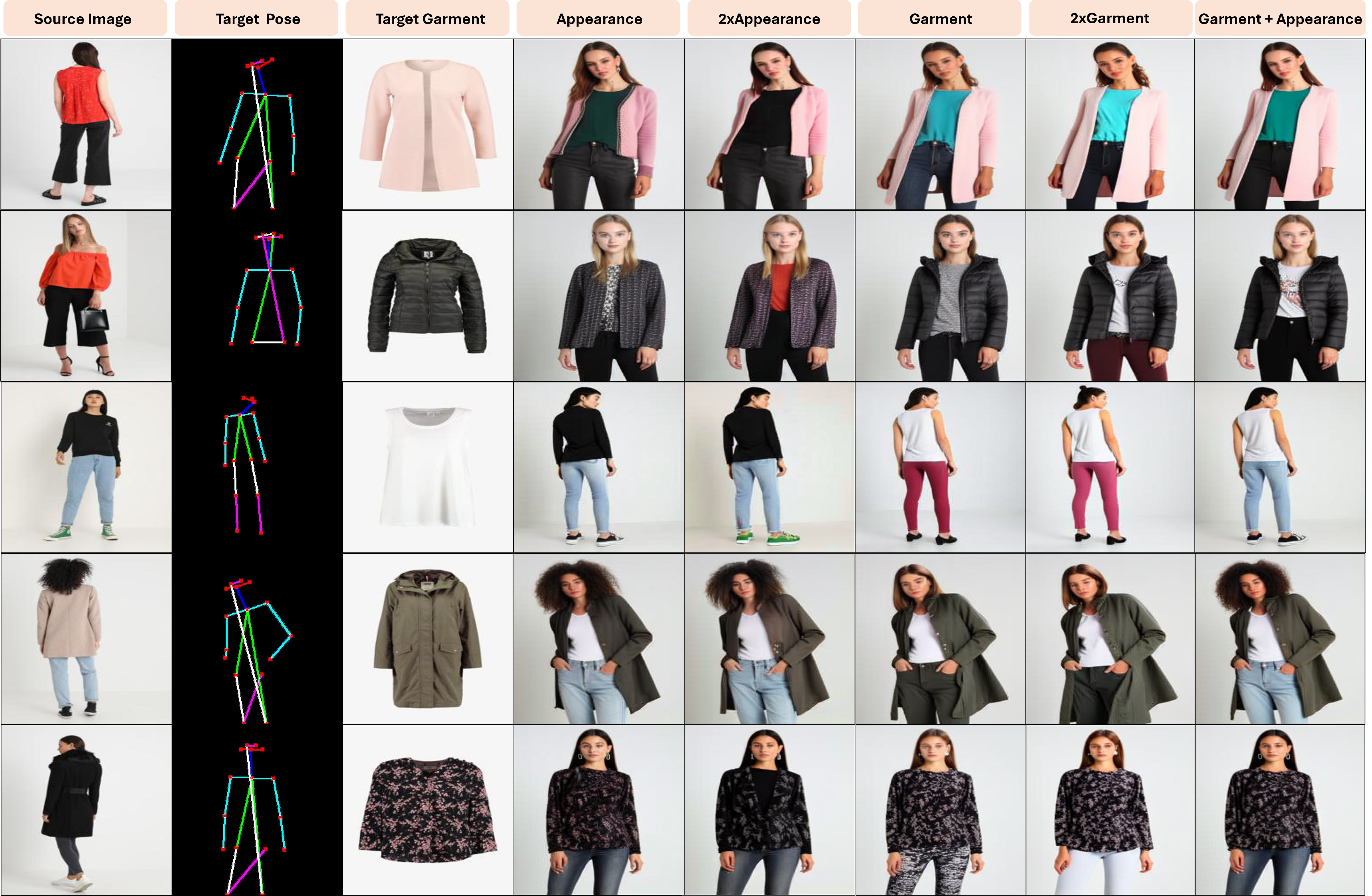}
\caption{The results of five experiments demonstrating the impact of different input variations to the appearance encoder. The experiments include (i) appearance only, (ii) 2x appearance, (iii) garment only, (iv) 2x garment, and (v) appearance + garment. The figure highlights how additional bias in queries improves the focus of the features, with the best results achieved by balancing both appearance and garment inputs.}
\label{fig: ablation1}
\end{figure*}

\begin{figure*}[t]
\centering
\includegraphics[width=\textwidth]{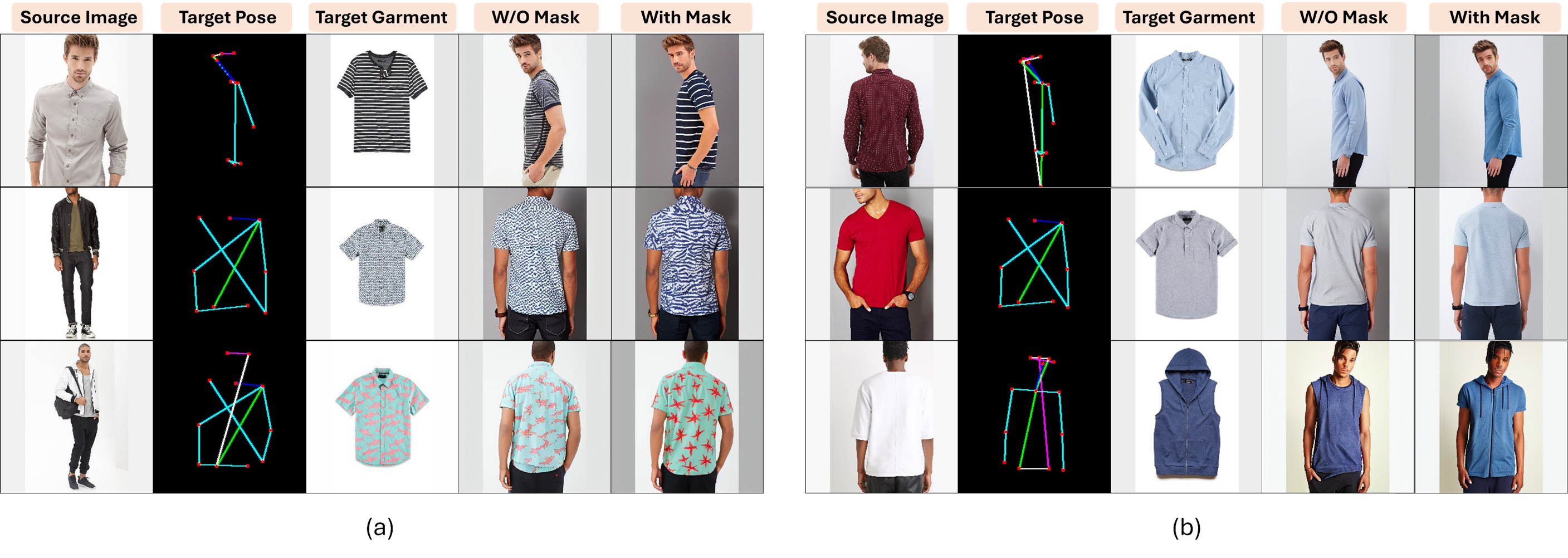}
\caption{Impact of gray masking on upper body segmentation, showing how masking removes garment details and contaminates appearance information, leading to distorted color and patterns in the output due to the cross-attention mechanism using biased queries from the source and garment images.}
\label{fig: ablation2}
\end{figure*}

\subsection{Ablation Study}

\subsubsection{Appearance Encoder Bias}
We demonstrate the effectiveness of our method, which adds two bias augmented queries—one from the source human image and one from the garment image—by conducting experiments with different input variations to the appearance encoder. We conducted five experiments: i) appearance only, ii) 2x appearance, iii) garment only, iv) 2x garment, and v) appearance + garment. Figure 4 demonstrates that the additional bias in the queries enhances feature focus. The appearance-only case captures appearance details more effectively than the garment-only case, but garment features merely reflected.

In particular, the analysis of the third row in Fig \ref{fig: ablation1} shows that when only appearance bias is added, the garment feature is almost excluded. Doubling the appearance bias (2x appearance) leads to improved preservation of the source image’s appearance, including finer details such as shoe color, which are not as well captured with a standard appearance bias. Conversely, when only garment bias is applied, the target garment is accurately represented, but other features, such as the original pants color, are not preserved.

Scaling the garment bias by 2 ensures that the target garment is well-represented, but still leads to the loss of other key features, such as the hairstyle. In contrast, applying both appearance and garment biases results in a balanced outcome where the target garment is accurately reflected while preserving non-garment features (e.g., pants color, hair color, and skin tone), which are critical for achieving realistic VTON results.

This experiment demonstrates that the latent space can be effectively controlled to balance the representation of both garment and appearance features. By adjusting the garment and appearance biases, the method maintains the necessary garment modifications while preserving the source appearance, resulting in optimal visual fidelity.

\subsubsection{Explicit Garment Synthesis}
In Fig \ref{fig: ablation2}, we present the results of an experiment comparing the effects of applying gray masking to the upper body region of source images against our method without masking.
In Figure 5(a), we observed that for garments with complex textures or patterns, such as striped clothing, applying the gray mask led to a simplified version of the patterns, resulting in fewer stripes or less intricate designs.
Additionally, as shown in Figure 5(b), we also noticed slight color distortions in the masked images.

Through our evaluation, we found that the approach without gray masking produced better results. This is because when gray masking is applied to the upper body region of the source image, it removes the detailed information. This not only loses the detailed garment information like color and patterns, it also contaminates the appearance information. This happens because our attention mechanism in the image generation process computes cross-attention with two queries from source human and garment images as bias. When masking is applied, the gray mask which takes the majority of the appearance information makes the model extract unrelated information. This hinders the denoising process and distorts the color and patterns of the output image. Additionally, the similarity between the gray mask and the background causes further confusion for the model, hampering the learning process and reducing output quality.

\section{Conclusion and Future Works}
In this work, we introduce Outfitting Diffusion with Pose-Guided Conditioning (ODPG), a novel virtual try-on (VTON) framework that seamlessly integrates garment, pose, and appearance features into a diffusion-based U-Net architecture. By leveraging a bias-augmented learnable query mechanism, our model effectively generates realistic VTON images without the need for explicit garment warping. It captures fine-grained details such as garment textures, fit, and natural alignment with dynamic poses. Our approach simplifies the virtual try-on pipeline by eliminating multi-stage processes, offering an efficient end-to-end solution that maintains high visual quality. Our experimental results demonstrate the model’s ability to generalize beyond the need for fully paired datasets, reducing dependency on exhaustive data requirements while still achieving impressive performance in generating realistic try-on results.

We are considering several promising directions for future research. One key area is extending the model to handle video inputs, enabling continuous and dynamic garment and pose transitions in video-based virtual try-on scenarios. This capability could provide more interactive and engaging user experiences. Additionally, the implicit learning strategy proposed in our model might be adapted to other domains that face data limitations or dependency challenges. By effectively addressing data scarcity, our method has the potential to be applied in various fields where obtaining extensive paired datasets is difficult. These future directions highlight the versatility and potential of our model for broader applications in both virtual try-on and other data-driven tasks.

\section*{Acknowledgement}
This study was conducted with support from the Creative Challenger Program (CCP) activity funding provided by the Teaching and Learning Enhancement Team at Korea University in 2024.

\bibliographystyle{apalike}
{\small
\bibliography{references}}

\end{document}